\documentclass[10pt,twocolumn,letterpaper]{article}

\usepackage{3dv}
\usepackage{times}
\usepackage{epsfig}
\usepackage{graphicx}
\usepackage{amsmath}
\usepackage{amssymb}
\usepackage{wrapfig}
\usepackage{multirow}
\usepackage{scalefnt}

\usepackage{url}
\usepackage{cite}
\usepackage{mathtools}
\usepackage{caption}

\usepackage[pagebackref=true,breaklinks=true,colorlinks,bookmarks=false]{hyperref}

\threedvfinalcopy 

\def\threedvPaperID{277} 

\AtBeginDocument{
  \abovedisplayskip     =0.1\abovedisplayskip
  \abovedisplayshortskip=0.1\abovedisplayshortskip
  \belowdisplayskip     =1.0\belowdisplayskip
  \belowdisplayshortskip=1.0\belowdisplayshortskip}

\ifthreedvfinal\fi
\begin{document}

\title{Distortion-Aware Self-Supervised 360\textdegree~Depth Estimation from \\ A Single Equirectangular Projection Image}

\author{Yuya Hasegawa\\
The University of Tokyo\\
Tokyo, Japan\\
{\tt\small hasegawa@hal.t.u-tokyo.ac.jp}
\and
Ikehata Satoshi\\
National Institute of Informatics\\
Tokyo, Japan\\
{\tt\small sikehata@nii.ac.jp}
\and
Kiyoharu Aizawa\\
The University of Tokyo\\
Tokyo, Japan\\
{\tt\small aizawa@hal.t.u-tokyo.ac.jp}
}

\maketitle

\begin{abstract}
360\textdegree~images are widely available over the last few years. 
This paper proposes a new technique for single 360\textdegree~image depth prediction under open environments. Depth prediction from a 360\textdegree~single image is not easy for two reasons. 
One is the limitation of supervision datasets -- the currently available dataset is limited to indoor scenes.
The other is the problems caused by Equirectangular Projection Format (ERP), commonly used for 360\textdegree~images, 
that are coordinate and distortion. 
There is only one method existing that uses cube map projection to produce six perspective images and apply self-supervised learning using motion pictures for perspective depth prediction to deal with these problems.
Different from the existing method, we directly use the ERP format. 
We propose a framework of direct use of ERP with coordinate conversion of correspondences and distortion-aware upsampling module to deal with the ERP related problems and extend a self-supervised learning method for open environments.
For the experiments, we firstly built a dataset for the evaluation, and 
quantitatively evaluate the depth prediction in outdoor scenes. 
We show that it outperforms the state-of-the-art technique.
\end{abstract}

\section{Introduction}
Thanks to the recent proliferation of consumer-level 360\textdegree~cameras, the use of 360\textdegree~images has become commonplace.
360\textdegree~cameras consist of multiple well-calibrated cameras and  can produce beautifully stitched 360\textdegree~images without any technical knowledge.
With the wider use of 360\textdegree~images, they are also attracting attention as a research target for computer vision, e.g. object detection\cite{8546070, Coors_2018_ECCV, Zhao_2020_AAAI}, super resolution\cite{9155477, Sevom2018360PS, 8901764} and scene semantic segmentation\cite{Jiang_ICLR_2019, Zhang_2019_ICCV}.
Acquiring depth information from a 360\textdegree~image is also an important task for autonomous systems such as autonomous driving cars or robots to understand and communicate with the world around them. While the depth information could also be acquired either from any active sensors such as LiDAR or passive stereo sensors, there is a growing demand for systems that use only inexpensive digital cameras whose number of devices is as small as possible. Therefore, using 360\textdegree~cameras which cover the large field-of-view (FoV) is an attractive choice in many real world scenarios.
\begin{figure}[t]
\begin{center}
\includegraphics[width=\linewidth, bb= 0 0 352 188]{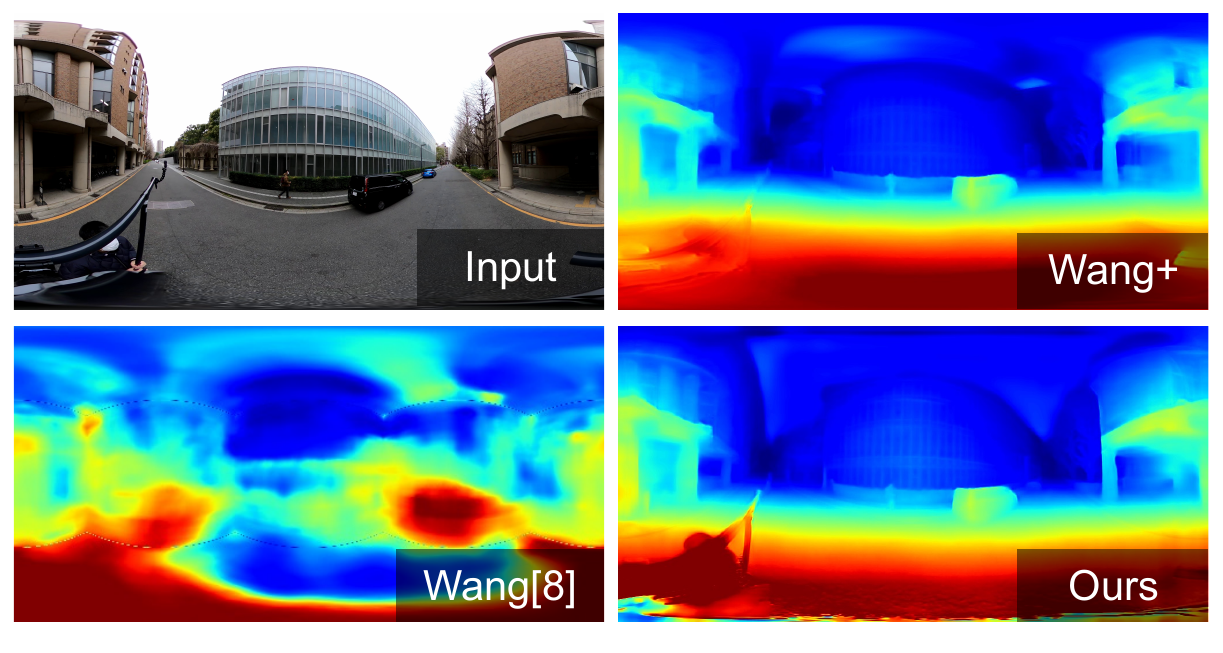}
\caption{{\bf Depth from a single panorama image.} Our self-supervised model produces sharper, and higher quality depth maps than existing work and its extension.}
\label{fig:top}
\end{center}
\end{figure}
The depth estimation based on monocular images has progressed rapidly thanks to the development of the data-driven methods, in particular, deep neural networks trained on large RGB-D datasets\cite{Eigen_2015_ICCV, laina2016deeper, Fu_2018_CVPR}. However, unlike methods for perspective images, ones using 360\textdegree~cameras are little or, if any, applicable to very limited scenes due to the lack of available annotated large training dataset.

To tackle this issue, Wang~\etal~\cite{Wang_2018_ACCV} have recently extended SfMLearner~\cite{Zhou_2017_CVPR}, which is the self-supervised single image depth estimation method designed for cameras with normal field-of-view, to the 360\textdegree~single image depth prediction task. Concretely, Wang~\etal projected 360\textdegree~videos in an Equirectangular projection (ERP) format into cubemaps which consist of six faces with 90\textdegree~FoV to avoid the spherical image distortion. Once depth maps for each face are individually predicted via convolutional neural networks (CNN), all the information is merged via the photometric consistency loss. While effective, their method has some obvious drawbacks. First, their divide-and-merge strategy (\ie, predicting depth maps from individual partial cubemaps and then merge them to get the final result) inevitably limits the network to take into account the global context in the entire 360\textdegree~view. Second, the depth maps predicted by their method are generally very blurry because the network can easily be overfitted to the texture-less region such as floor and ceiling or sky and ground which generally appear in high-latitude areas in 360\textdegree~images.

In this paper, we present a novel self-supervised deep-learning-based algorithm to predict the depth map from a single 360\textdegree~image. Our work is inspired by SfMLearner~\cite{Zhou_2017_CVPR} and its 360\textdegree~extension~\cite{Wang_2018_ACCV}, however we don't decompose the spherical image into multiple normal FoV images but instead directly applying CNN to an ERP image to maximally benefit from the entire scene context.  Since directly applying CNN to ERP images is known to have an issue of incorrect prediction near the poles due to the distortion caused by the sphere-to-plane projection~\cite{Coors_2018_ECCV, Zioulis_2018_ECCV}, we propose {\it Distortion-aware Upsampling Module} that weights the feature map of different layers based on the amount of the distortion calculated analytically. Not only does it make the network robust against the sphere-to-plane distortion, but it also prevents the network from overfitting to observations around high-latitude areas with relatively little texture, resulting in sharper depth maps. Since self-supervised learning is more effective in scenarios where it is difficult to acquire annotated training data. Therefore, our work focuses more on the challenging outdoor scenes rather than indoor scenes where synthetic datasets are often available. To this end, we train our model on large-scale outdoor walk-around videos~\cite{Sugimoto_2020} with no depth annotations and evaluate it on our own outdoor 360\textdegree~image dataset where ground truth depth maps are provided using the multi-view stereo (MVS) algorithm since there is no publicly available outdoor 360\textdegree~RGB-D image dataset. We compare our method directly against Wang~\etal~\cite{Wang_2018_ACCV} and the extension of it where the backbone was upgraded to a more modern network architecture for the fair comparison.

Our contributions are summarized as follow,
\begin{itemize}
    \item (1) We present the self-supervised single 360\textdegree~image depth estimation method that directly handles an ERP format without relying on the cubemap projection.
    \item (2) We propose the Distortion Aware Upsampling Module which makes the network robust to the distortion caused by the sphere-to-plane projection and prevent our model from over-fitting to the high-latitude texture-less observations.
    \item (3) We present the first 360\textdegree~RGB-D image evaluation benchmark dataset where the ground truth annotations are sparsely provided based on the MVS algorithm.
    \item (4) We present that our method both quantitatively and qualitatively outperformed~\cite{Wang_2018_ACCV} and its extension with better backbone architecture.
\end{itemize}

\begin{figure}[t]
\begin{center}
\includegraphics[width=\linewidth, bb= 0 0 728 418]{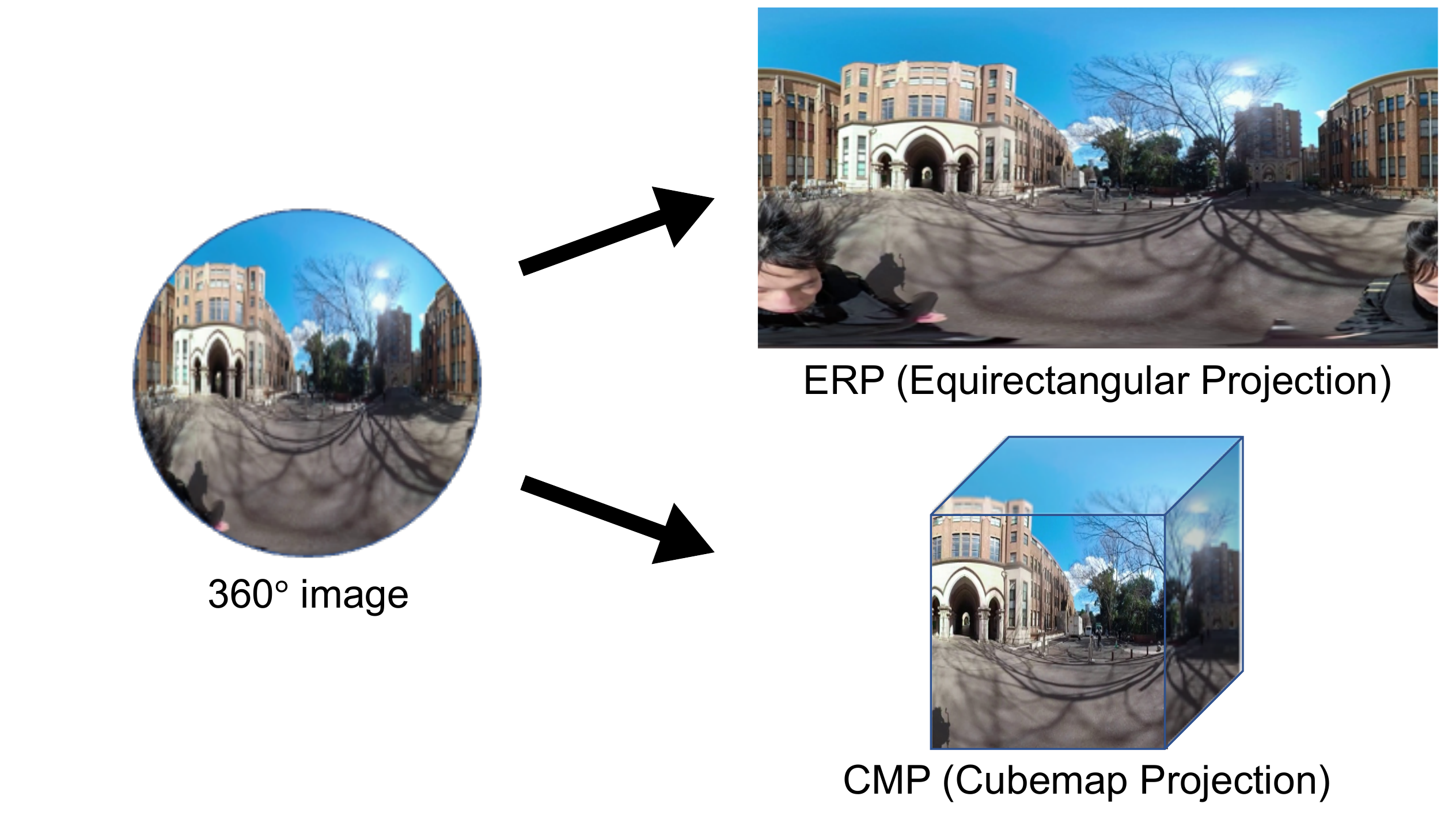}
\caption{{\bf Commonly used 360 \textdegree~image expression methods.} ERP includes distortions but can be represented by a single image. CMP splits a 360\textdegree~image into six individual images, but can be represented without distortions.}
\label{fig:projection}
\end{center}
\end{figure}

\begin{figure*}[th]
\begin{center}
\includegraphics[width=\linewidth, bb= 0 0 892 294]{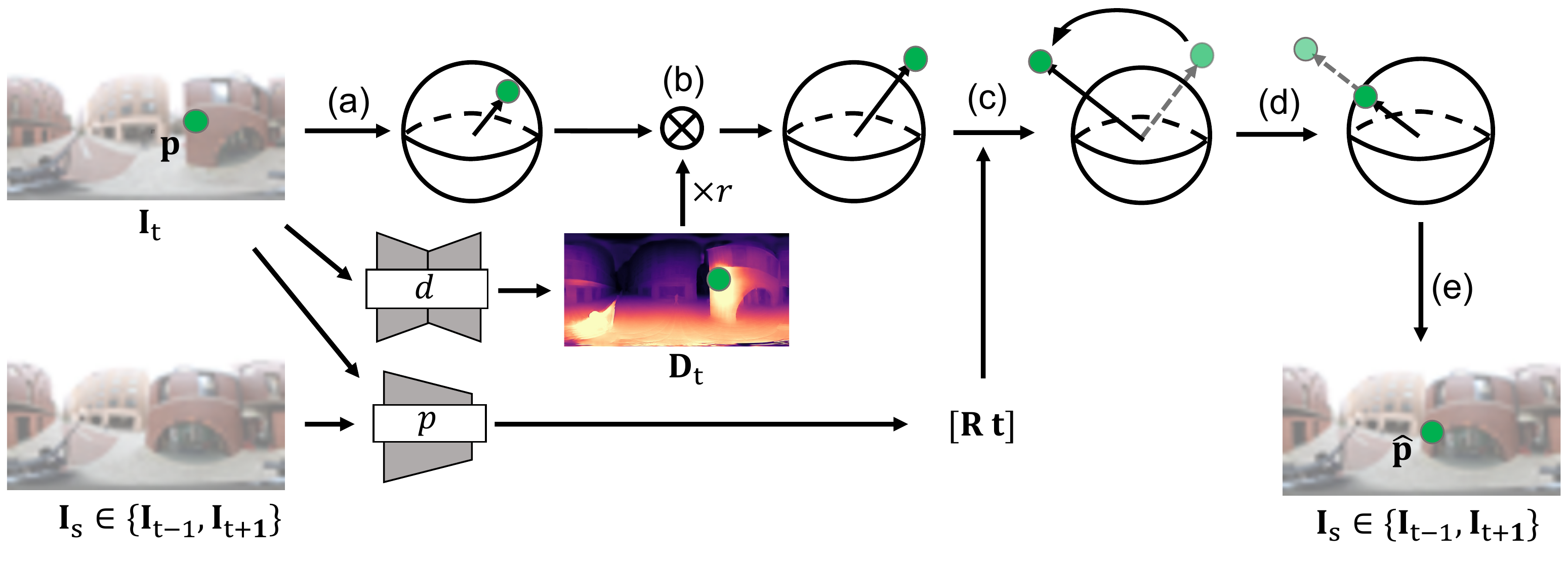}
\caption{{\bf Overview of 360\textdegree~depth and pose estimation.} In training phase, we use two 360\textdegree~images $\mathbf{I_t}$ and $\mathbf{I_s}$ as our input. The depth predictor ($d$) estimates depth of $\mathbf{I_t}$ and the pose predictor ($p$) predicts relative pose between $\mathbf{I_t}$ and $\mathbf{I_s}$. Then, using the $\mathbf{D_t}$ and the estimated relative pose, project each point in $\mathbf{I_t}$ to $\mathbf{I_s}$. The parameters of networks are updated to decrease the photometric reconstruction error based on this projection.}
\label{fig:sfmlearner}
\end{center}
\end{figure*}

\section{Related Work}
\noindent \textbf{Monocular Depth Estimation from Normal FoV images}

Unlike the stereo matching or active sensing, depth estimation from a single image inherits ambiguity where infinite number of 3-D scenes could result in the same 2-D image. The main challenges underlying this problem is how to disambiguate the scene as human visual systems do. Therefore, the data-driven approach which can capture the high-level scene context was promising from the beginning of this task unlike other 3-D reconstruction problems. Saxena~\etal~\cite{Saxena2006} presented the first learning-based algorithm that tried to estimate a full metric depth map from a single image. In this work, the input image was decomposed into patches and a single depth value is predicted for each patch by the discriminatively-trained Markov Random Field (MRF) which was trained on the 3-D laser scanner data. The assumption that a 3-D scene can be expressed with small planar surfaces were preferred by researches for a decade because it reduces the computational cost and improves robustness to outliers\cite{4531745}.

In 2014, Eigen~\etal~\cite{10.5555/2969033.2969091} introduced the first deep convolutional neural networks for the single image depth estimation task and achieved better performance than conventional algorithms. They proposed the combinatorial architecture of coarse and fine sub-modules which allowed the network to capture both global context and local details. Inspired by this work, many supervised deep-learning based methods have emerged and made great progress in this problem\cite{Eigen_2015_ICCV, laina2016deeper, Fu_2018_CVPR}. These methods basically require densely and accurately annotated RGB-D images. However, in most cases, obtaining such information is labor-intensive or difficult. In such cases, synthetic data is often used, but according to \cite{Godard_2019_ICCV}, it is not trivial to reproduce the complex environment of the real world. Therefore, some self-supervised learning based methods that do not require annotated data have been proposed in recent years. Zhou~\etal~\cite{Zhou_2017_CVPR} presented the pioneer work named SfMLearner which is a self-supervised learning framework consisting of monocular depth and camera motion estimation networks from unstructured video sequences. The output of the networks are bound via the loss based on warping nearby views to the target using the computed depth and pose as common structure-from-motion algorithms do. Inspired by this work, GeoNet~\cite{Qi_2018_CVPR} proposed a self-supervised learning framework for the joint estimation of depth, camera pose and optical flow. While this improved the optical flow estimation, the depth prediction accuracy didn't benefit from the joint prediction. The multi-task strategy was further investigated where the depth and pose estimation was combined with the prediction of the surface normal~\cite{Zhang_2019_CVPR}, edges~\cite{Xu_2018_CVPR} for jointly improving the performance of individual tasks.

More recently, Godard~\etal~\cite{Godard_2019_ICCV} improved the self-supervised monocular depth estimation network by introducing new architecture and loss functions which are a minimum reprojection loss to handle occlusions, multi-scale sampling strategy to reduce artifacts, and the auto-masking loss to ignore training pixels that violate camera motion assumptions. These components contributed to yield state-of-the-art monocular self-supervised depth estimation on the KITTI dataset and became the current de facto standard algorithm. Our work extends this method to tackle the monocular depth prediction from a single {\it 360\textdegree}~image.
\\\\
\noindent \textbf{Monocular Depth Estimation from 360\textdegree~images}

While the annotated data acquisition is a big challenge in training CNN for the single image 360\textdegree~depth estimation, it is also non-trivial to apply convolutions to 360\textdegree~images because the observation is stored on a sphere rather than a plane. The most straight-forward way to tackle this challenge is to project the spherical coordinates into planar coordinates, then apply the CNN to the resulting 2-D image (\eg, ERP or cylindrical projection~\cite{Sharma2019UnsupervisedLO}). In this manner, Zioulis~\etal~\cite{Zioulis_2018_ECCV} presented the first U-Net like dense 360\textdegree~depth estimation network trained on the synthetic 360\textdegree~dataset. Since any sphere-to-plane projection introduces distortion and making the resulting convolutions inaccurate, they concatenated the filter output of different kernel widths reflecting the amount of distortion that varies by latitude. While effective, their datasets only cover indoor cases, limiting the networks' applicability to outdoor settings. In addition to that, heuristically concatenating filter outputs of different kernel widths often generate blurry results even thought it is globally consistent. 

\begin{figure}[t]
\begin{center}
\includegraphics[width=\linewidth, bb=0 0 354 189]{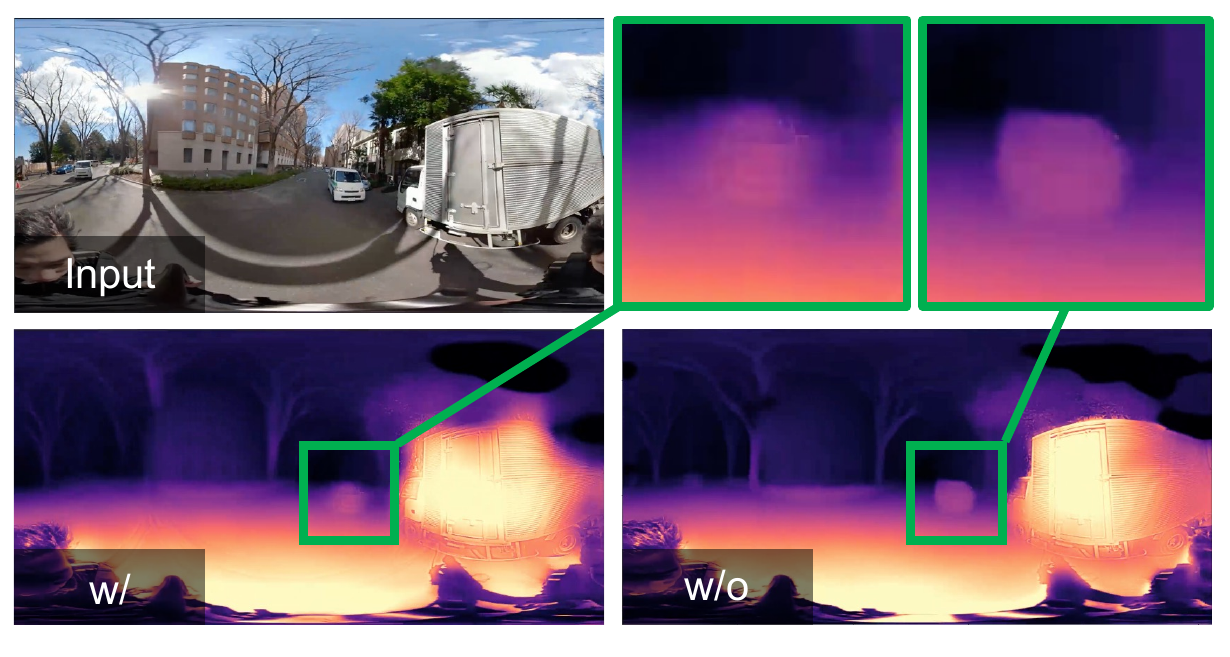}
\caption{{\bf Results of a preliminary experiment of the preprocessing
blocks proposed by Zioulis~\etal~\cite{Zioulis_2018_ECCV}.} The boundary is blurred and the details are not captured. The method of changing the shape of convolution kernels provides overall consistency, but blurs the boundaries.  }
\label{fig:preliminary}
\end{center}
\end{figure}


Another strategy of applying CNN to spherical data is called CubeMap Projection(CMP) where a 360\textdegree~image is repeatedly project to tangent planes around the sphere, each of which is then fed to the CNN. A self-supervised learning method for the monocular 360\textdegree~depth estimation based on CMP has been proposed by Wang~\etal~\cite{Wang_2018_ACCV}. This work follows SfMLearner~\cite{Zhou_2017_CVPR} where both depth maps of every faces and pose of cameras were jointly predicted in an end-to-end manner while the consistency of different faces were encouraged by the photometric consistency loss. This approach is attractive because the annotated trained data is unnecessary and no projective distortion is observed in the cubemap. However, as have already been dscribed, their divide-and-merge strategy inevitably limits the network to take into account the global context in the entire 360\textdegree~view and the depth maps predicted by their method are generally very blurry because the network can easily be overfitted to the texture-less region. Alisha~\cite{Sharma2019UnsupervisedLO} has also proposed the method for self-supervised learning of depth and ego-motion from a single 360\textdegree image but used the  panoramic projection instead of CMP. While  projection is continuous and avoids increasing distortion toward the poles, it cannot represent the entire sphere.

In our work, we present the self-supervised learning based framework for the single 360\textdegree~depth estimation inspired by SfMLearner~\cite{Zhou_2017_CVPR} in similar with~\cite{Wang_2018_ACCV, Sharma2019UnsupervisedLO}, but we can directly input the ERP image (\ie, cover the whole $360\times 180$ FoV) rather than cube maps or  projection image by introducing distortion-aware upsampling module that naturally handles different amount of distortions that varies by latitude. 
\section{Method}



In this section, we introduce our self-supervised learning framework for the monocular 360\textdegree~depth estimation from a single ERP image. The framework is illustrated in Fig.\ref{fig:sfmlearner}. Our work basically extends the SfMLearner framework~\cite{Zhou_2017_CVPR} and its extension~\cite{Godard_2019_ICCV} to take a 360\textdegree~image in ERP format as input while preventing artifacts caused by distortions due to projection from spherical to planar surfaces. Here, we firstly formulate the geometric transformation between two different ERP images via a coordinate transformation method (Sec.\ref{sec:coordinate}), then we introduce the distortion-aware upsampling module that accounts for different amount of distortions that varies by latitude (Sec.\ref{sec:DAUM}).

\subsection{ERP Specific Coordinate Transformation}
\label{sec:coordinate}
Following the basic SfmLearner architecture~\cite{Zhou_2017_CVPR}, our network consists of depth and relative pose predictors. Similar to the normal perspective cameras, the pose of a 360\textdegree~camera could also be represented by its translation and rotation in world coordinate system. Given the correct depth map of a 360\textdegree~view and the relative camera rotation/translation between two different 360\textdegree~views, the observation from one view could be projected to the other if there is no occlusions in the scene. Given two adjacent frames from a 360\textdegree~video stream, the self-supervised learning framework proceeds by updating the parameters of depth and pose predictors so that the errors between projected image and the original image decrease.


Concretely, we denote input target and source ERP images as $\mathbf{I_t}\in\mathbb{R}^{H\times W\times 3}$ and $\mathbf{I_s}\in \{\mathbf{I}_{t-1}, \mathbf{I}_{t+1}\}$ where $t$ is the target video frame index. We also denote the output of target depth predictor as $\mathbf{D_t}\in\mathbb{R}^{H\times W\times 1}$ and express the relative pose between $\mathbf{I_t}$ and $\mathbf{I_s}$ as $\mathbf{R}\in SO(3)$ and $\mathbf{t}\in\mathbb{R}^3$ respectively. The main process in the self-supervised learning framework requires mapping from $\mathbf{I_t}$ to $\mathbf{I_s}$ based on the predicted depth and pose information. Since $\mathbf{I_t}$ and $\mathbf{I_s}$ are ERP format where images are projected from the unit sphere, the geometric structure is different from that of perspective projection camera models, and the coordinate transformation in~\cite{Zhou_2017_CVPR} cannot be directly applied to. Therefore, we propose a method that can be applied to ERP images by using a transformation between polar and Cartesian coordinates.

Assuming polar coordinates $\mathbf{p}\triangleq (\theta_{\mathbf{p}}, \phi_{\mathbf{p}})$ and $\hat{\mathbf{p}}\triangleq (\theta_{\hat{\mathbf{p}}}, \phi_{\hat{\mathbf{p}}})$ in $\mathbf{I_t}$ and $\mathbf{I_s}$ sees the same 3-D point, the mapping from $\mathbf{p}$ to $\hat{\mathbf{p}}$ is represented as follow.
\begin{itemize}
\item (a) Transform the polar coordinate representation $(\theta_{\mathbf{p}}, \phi_{\mathbf{p}})$ to the Cartesian coordinate representation $(x_{\mathbf{p}}, y_{\mathbf{p}}, z_{\mathbf{p}})$ on a unit sphere.
\item (b) The depth value $\mathbf{D_t} (\mathbf{p})$ is multiplied by the Cartesian coordinates to get $P = (d_{\mathbf{p}}x_{\mathbf{p}}, d_{\mathbf{p}}y_{\mathbf{p}}, d_{\mathbf{p}}z_{\mathbf{p}})$. 
\item (c) Apply affine transformations to the coordinates obtained in (b) using $[\mathbf{R}, \mathbf{t}]$ to get $\hat{P} = (\hat{x}_{\mathbf{p}}, \hat{y}_{\mathbf{p}}, \hat{z}_{\mathbf{p}})$. 
\item(d) Projecting the transformed coordinate onto the unit sphere of the source view by normalizing the representation by $\hat{r}$, where $\hat{r}=\sqrt{\hat{x}_{\mathbf{p}}^2+\hat{y}_{\mathbf{p}}^2+\hat{z}_{\mathbf{p}}^2}$. 
\item (e) By using the polar coordinate representation, the reprojected coordinates of $\mathbf{\hat{p}}$ can be obtained as follow.
\end{itemize}

\begin{align}
\label{equation:1}
    \large
    \mathbf{P}=\left[\begin{array}{c}x_\mathbf{p} \\y_\mathbf{p} \\z_\mathbf{p}\end{array}\right] &= \left[\begin{array}{c}d_{\mathbf{p}}\cos\theta_\mathbf{p}\sin\phi_\mathbf{p} \\d_\mathbf{p}\sin\theta_\mathbf{p} \\ d_\mathbf{p}\cos\theta_\mathbf{p}\cos\phi_\mathbf{p}\end{array}\right] \\
\label{equation:2}
\mathbf{\hat{P}}=\left[\begin{array}{c}\hat{x}_\mathbf{p} \\\hat{y}_\mathbf{p} \\\hat{z}_\mathbf{p}\\1\end{array}\right]&=\left[\begin{array}{cc}\mathbf{R} & \mathbf{t}\end{array}\right]\left[\begin{array}{c}x_\mathbf{p}\\y_\mathbf{p} \\z_\mathbf{p} \\1\end{array}\right] \\
\label{equation:3}
\ \mathbf{\hat{p}}= \left[\begin{array}{c}\hat{\theta}_\mathbf{p} \\\hat{\phi}_\mathbf{p}\end{array}\right]&=\left[\begin{array}{c}\arctan \bigl(\frac{\hat{x}_\mathbf{p}}{\hat{r}}\bigr)\\\arctan \bigl(\frac{\hat{y}_\mathbf{p}}{\hat{r}}\bigr)\end{array}\right]
\normalsize
\end{align}

\begin{figure*}[t]
\begin{center}
\includegraphics[width=\linewidth, bb=0 0 705 239]{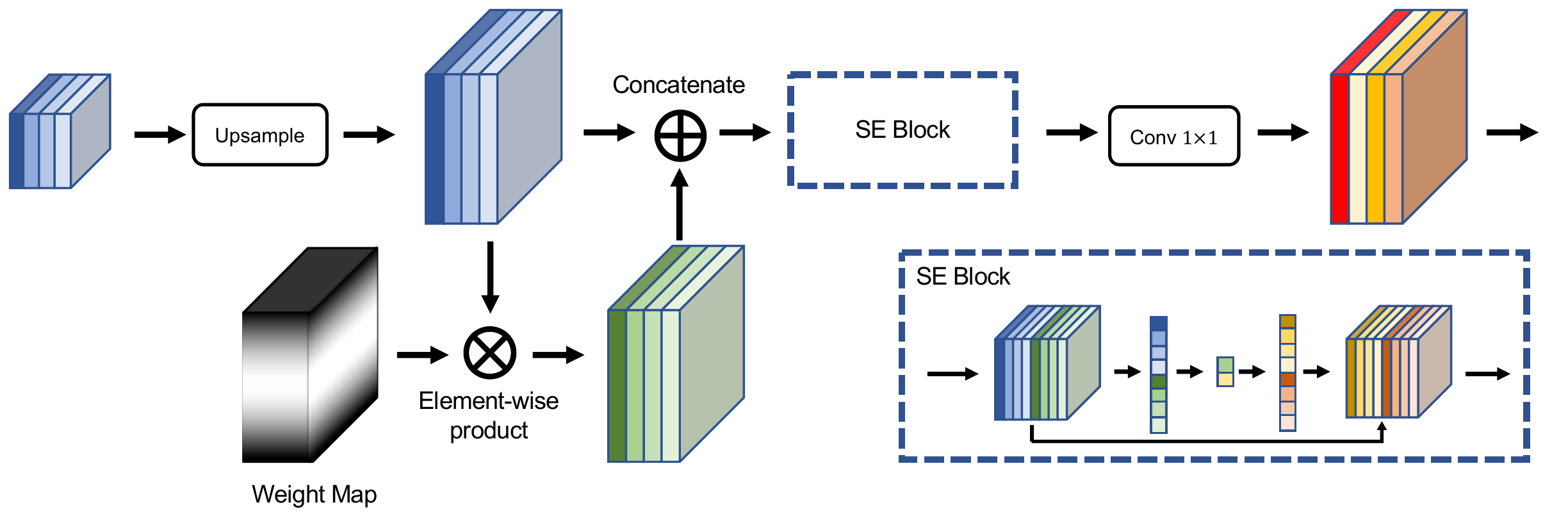}
\caption{{\bf Distortion Aware Upsampling Module.} Given the feature maps from the previous layer, Distortion Aware Upsampling Module concatenates the original feature map and the one weighted by the amount of the distortion caused by the sphere-to-plane projection and then feed them to the SE Block for the adaptive feature selection. By introducing this module, the sparse and contaminating high-latitude information is suppressed in feature map upsampling steps.}
\label{fig:DASE}
\end{center}
\end{figure*}

\subsection{Distortion\,Aware\, Upsampling\, Module}
\label{sec:DAUM}


In the ERP image, the distortion of the image becomes stronger at higher latitude areas. Specifically, the spherical coordinate is stretched in the horizontal direction on the plane. This means that the semantic information of the content in the image also becomes more sparse as it locates away from the equator. In addition, the existing consumer 360\textdegree~camera captures images aligned with the gravity direction, therefore the high-latitude regions basically represent less-textured areas such as sky and ground. This becomes a problem when using U-Net architecture\cite{RFB15a}, which is generally used in depth estimation models. The decoder module of U-Net is designed to convey semantic information from deeper layers, but this design leads to erroneous inferences away from the equator because of the over-fitting to the distorted, less-textured high-latitude areas. In practice, our depth model infers depth maps with different resolutions from each layer at different depths and we empirically found the estimated depth map from the deeper layer is more error-prone. To make matters worse, features from deeper layers are observed to have a stronger influence on the final output than the skip connection.
Based on this observation, we propose to replace the basic upsampling layer in the depth estimation model by  our Distortion Aware Upsampling Module to discard contaminating information at high latitude areas. The module is illustrated in Fig.\ref{fig:DASE}. Given the filter output from the previous layer, we concatenate the original feature map and weighted feature map whose weight is based on the amount of distortion calculated using the following equation as in Zhou~\etal~\cite{8652269}. \begin{equation}
W(u, v) = \cos\bigl(\bigl( v-\frac{H}{2}+\frac{1}{2}\bigr) \frac{\pi}{H} \bigr)
\end{equation}
Here $H$ is the height of the ERP image and $u,v$ are the coordinates in the ERP image. By applying our distortion-aware upsampling module, the contribution of features located away from the equator gradually decreases. The concatenated features are then passed to SE Block as proposed by Hu~\etal~\cite{Hu_2018_CVPR} to extract important features. Then, by applying 1x1 convolution, the number of channels is reduced and only the necessary features are propagated. This allows the network to dynamically weight features from deeper layers that are far from the equator. While the weight maps used in Distortion Aware Upsampling Module can prevent the network from being corrupted by the distortion in the ERP image, we find that putting this module to all the layers could generate blurry result because it may also lose local structure details of the high-latitude regions. Therefore, we don't use the weight maps in depth estimation models except for upsampling layers. For the pose inference model, we did not use the weight maps because the estimation is based on the information of the entire image, such that the top and bottom of the image are more sensitive to changes in angle.

\section{Dataset}
\subsection{Training Dataset}
In this study, we used 25 of 360\textdegree~videos for training and 5 videos for validation which were captured in the university campus \cite{Sugimoto_2020}. By sampling images from video, we used 8640 images for the training data and 1441 images for the validation data. Since the data used was walking video, the speed of the video was constant, so the pose inference model could not be trained well without randomness in the sampling period of the image. Therefore, we repeatedly sample every random number of frames between 5 and 30 frames from a video shot at about 30 fps. Because it is a data-dependent feature, the range of frames between samples were chosen empirically to ensure stable learning.

\subsection{Evaluation Dataset}

\begin{wrapfigure}{l}[0mm]{45mm}
  \centering
  \includegraphics[keepaspectratio, width=45mm, bb=0 0 407 308]{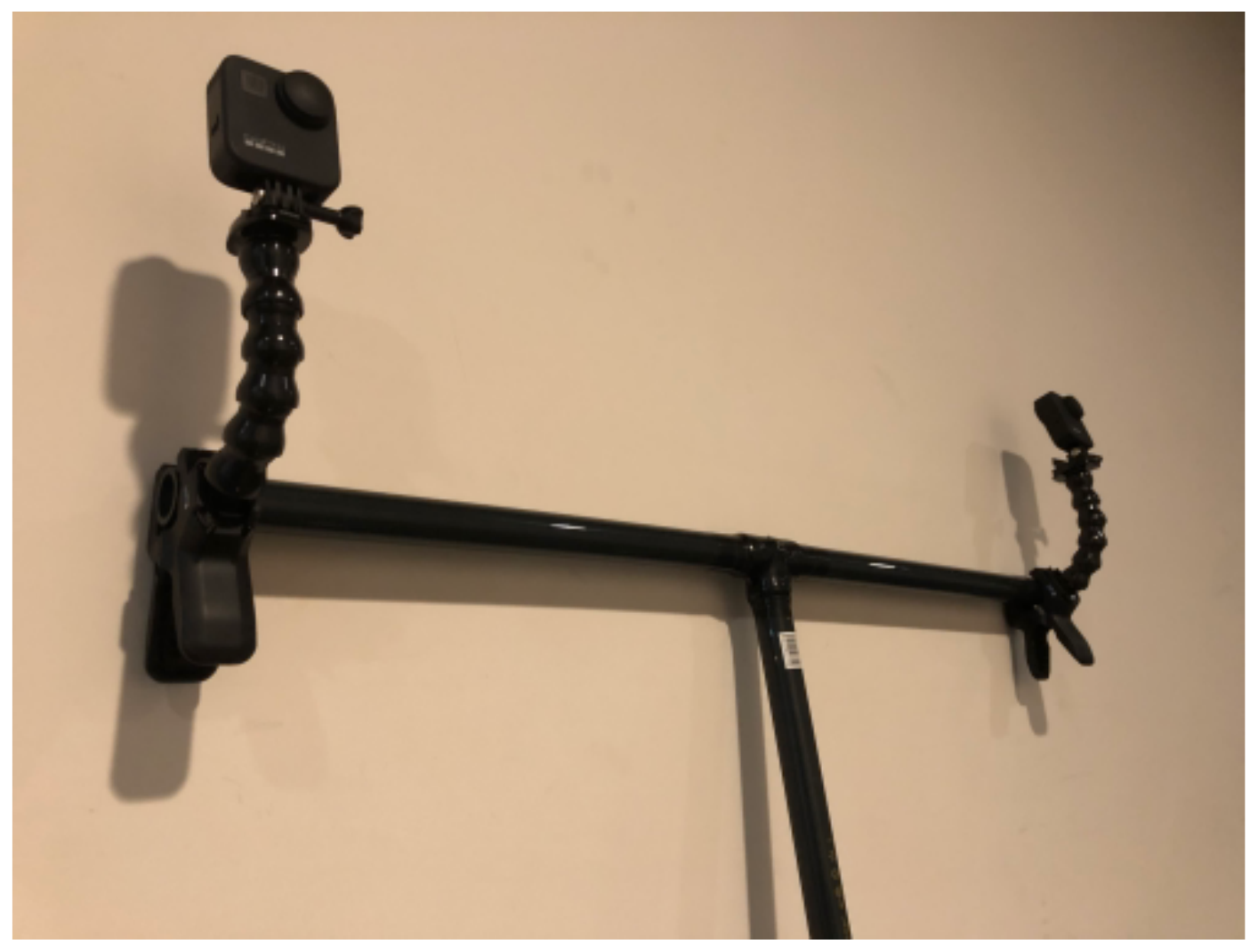}
  \caption{{\bf Camera rig.}}
  \label{fig:rig}
\end{wrapfigure}

We have built a dataset for evaluation in outdoor environments.
In this study, we use Multi-View Stereo (MVS) to built dataset for evaluation of monocular depth estimation as well as Li~\etal~\cite{Li_2019_CVPR} and Yoon~\etal~\cite{Yoon_2020_CVPR}.
Since a single camera cannot give absolute scale information to the depth data, two cameras (GoPro Max) are equipped on the rig at 1m intervals as shown in Fig.\ref{fig:rig}. From the distance between the two cameras, absolute scale information is given to the depth data.
In the dataset, we manually masked the noisy parts of the MVS, such as the photographer, so that the correct data would not contain the noise.
At each shooting point, the SfM and MVS algorithms in {\it Metashape}~\cite{metashape} were performed using about 70 images from two cameras combined within about 10-meter square. Finally, a dataset of 179 images was built and used for evaluation. Note that since the MVS generate less surface points on the smooth texture-less surfaces, we only used the points of high confidence for the evaluation (See details below). 
\section{Expreriments}
\subsection{Implementation Detail}
Our depth estimation network is an extension of the network proposed by Godard~\etal~\cite{Godard_2019_ICCV}. We use a ResNet50\cite{He_2016_CVPR} as our encoder and start with weights pretrained on ImageNet\cite{deng2009imagenet}. Our depth decoder uses sigmoid at the output. We convert the sigmoid output $\sigma$ to depth with $D = 1/(a\sigma+b)$, where a and b are chosen so that $D$ falls between 0.1 and 100.
For our pose estimation network, we follow \cite{Godard_2019_ICCV} and use ResNet50\cite{He_2016_CVPR} for encoder. The depth estimator and pose estimator use separate encoders.
The resolution of the input and output images used for training and evaluation is 512$\times$1024. 
We use horizontal flips with half the probability and apply the following augmentations, random brightness, contrast, saturation, and hue fitter with respective range of $\pm0.2, \pm0.2, \pm0.2$ and $\pm0.1$. 
Our models are implemented in PyTorch\cite{NEURIPS2019_bdbca288}. The models are trained with the Adam\cite{DBLP:journals/corr/KingmaB14}. The training is performed for a total of 20 epochs with the batch size of 6. The first 15 epochs have a learning rate of $10^{-4}$ , and the next 5 epochs have a learning rate of $10^{-5}$.

\subsection{Baseline}

We compared against CMP based method\cite{Wang_2018_ACCV}. However, as shown in Table 1, it is not a fair comparison because the backbone models of Wang~\etal~\cite{Wang_2018_ACCV} method and our method are different. Therefore, the backbone of Wang~\etal~\cite{Wang_2018_ACCV} method is aligned with our method as Wang+ for fairer comparison. Note that Wang~\etal~\cite{Wang_2018_ACCV} and Wang+ used the implementation reproduced by us in this study.

\ 

\begin{table}[h!]
\begin{center}
\caption{{\bf Backbone Models}}
\label{fig:backbone}
\begin{tabular}{lcc}
\hline
Backbone & Zhou~\etal~\cite{Zhou_2017_CVPR} & Godard~\etal~\cite{Godard_2019_ICCV} \\ \hline
Wang\cite{Wang_2018_ACCV}    & \checkmark  &               \\
Wang+                           &             &  \checkmark \\
Ours                            &             &  \checkmark  \\ \hline
\end{tabular}
\end{center}
\end{table}

\begin{table*}[t]
\scriptsize\centering
\caption{{\bf Quantitati results.}}
\scalebox{1.2}[1.2]{
\begin{tabular}{lcccccccc}
\hline
\multicolumn{1}{c}{\multirow{2}{*}{Method}} & \multicolumn{4}{c}{Lower the better} &  & \multicolumn{3}{c}{Higher the better}             \\ \cline{2-5} \cline{7-9} 
\multicolumn{1}{c}{}                        & Abs Rel  & Sq Rel & RMSE  & RMSE log &  & $\delta<1.25$ & $\delta<1.25^2$ & $\delta<1.25^3$ \\ \hline
Wang\cite{Wang_2018_ACCV}                                        & 0.578    & 5.455  & 9.776 & 0.711    &  & 0.303         & 0.549           & 0.713           \\

Wang+                                 & 0.288    & 4.604  & 4.346 & 0.263    &  & 0.734         & 0.917           & 0.965           \\ 

Ours                                   & \bf{0.236}    & \bf{2.450}  & \bf{4.240} & \bf{0.249}    &  & \bf{0.762}         & \bf{0.932}           & \bf{0.971}           \\ \hline
\end{tabular}
}
\label{table:result}
\end{table*}

\begin{table*}[t]\scriptsize\centering
\caption{{\bf Ablation of Distortion Aware Upsampling Module}. }
\label{table:DASE}
\scalebox{1.2}[1.2]{
\begin{tabular}{lcccccccc}
\hline
\multicolumn{1}{c}{\multirow{2}{*}{Method}} & \multicolumn{4}{c}{Lower the better} &  & \multicolumn{3}{c}{Highter the better}                                         \\ \cline{2-5} \cline{7-9} 
\multicolumn{1}{c}{}                        & Abs Rel  & Sq Rel & RMSE  & RMSE log &  & $\delta<1.25$ & $\delta<1.25^2$ & $\delta<1.25^3$ \\ \hline
w/o                                         & 0.249    & 2.721  & \bf{4.232} & 0.254    &  & 0.748                    & 0.927                    & 0.969                    \\
w/                                          & \bf{0.236}    & \bf{2.450}  & 4.240 & \bf{0.249}    &  & \bf{0.762}                    & \bf{0.932}                    & \bf{0.971}               \\ \hline
\end{tabular}
}
\end{table*}

\begin{figure*}[h!]
\begin{center}
\includegraphics[width=\linewidth, bb=0 0 872 209]{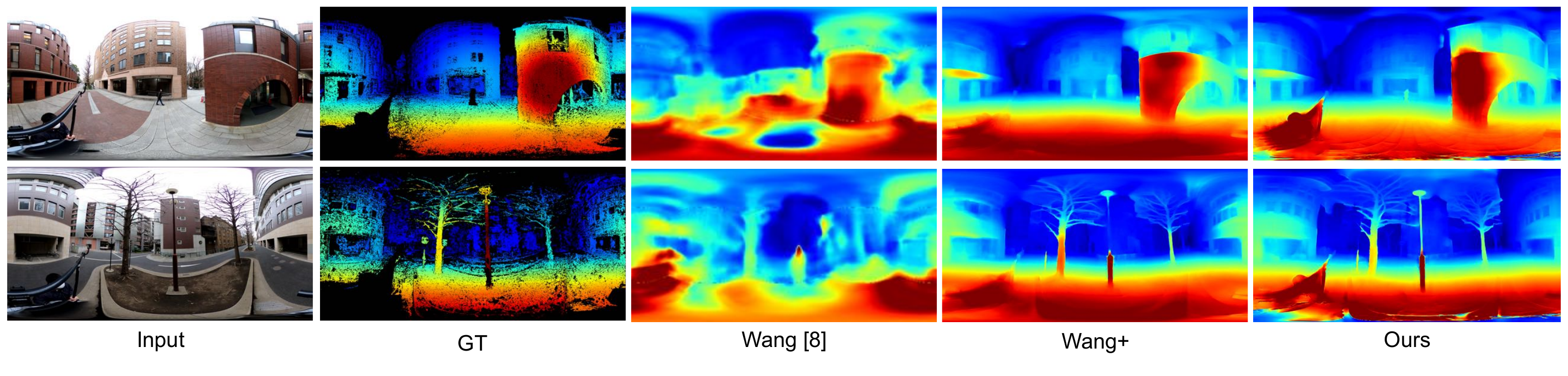}
\caption{{\bf Qualitative comparison of the 360\textdegree~depth estimation from each method.}}
\label{fig:result}
\end{center}
\end{figure*}

\subsection{Result}
Table \ref{table:result} shows the results of the quantitative evaluation. For more accurate evaluation, areas that empirically contain errors in the MVS, such as where the photographer is shown or areas with few textures, were manually masked. The proposed method outperformed the existing methods in all metrics except RMSE. 
Fig.\ref{fig:result} shows the output results for the evaluation dataset.
The qualitative results also show that the proposed method enables accurate estimation without blurring at the boundary of the object as compared with the existing method.
In the evaluation data, there is a large difference in the estimation results in the areas that are not included in the evaluation due to masks, etc. Therefore, it is considered that there is actually a greater difference in performance than shown in Table \ref{table:result}.

These results show that although the ERP representation contains distortions, deep learning models can be trained by redesigning the coordinate system so that the system for perspective projection can be extended to ERP, and incorporating modules to account for ERP-specific distortion such as the Distortion Aware Upsampling Module.
On the other hand, the CMP-based method can be directly applied by dividing the images into a group of undistorted images, but the error is considered to be larger than that of the ERP-based method because it requires estimation from only images with a limited viewing angle.

Table \ref{table:DASE} shows a comparison of the results with and without the Distortion Aware Upsampling Module. The effectiveness of not only extending the method of Godard~\etal~\cite{Godard_2019_ICCV} to a coordinate system suitable for 360° images, but also using the Distortion Aware Upsampling Module proposed in this study is demonstrated.

\section{Limitation and Future Work}
One of the future tasks is to expand the dataset for evaluation. In addition, it is necessary to create datasets that were taken under various environments for more accurate evaluation. Another issue is that the evaluation dataset does not include the areas where depth estimation is not performed by MVS. It is also necessary to include areas such as walls that don't have textures in the evaluation.
Considering that it will be used more widely in autonomous systems, it is necessary to perform verification in a wider variety of environments.For example, it can be used in an environment where people come and go. In this study, we created a dataset using MVS with two cameras, so that no moving objects are captured during the shooting. In an environment where people walk around, it is necessary to create a dataset that takes into account moving objects, so for example, it is possible to create a dataset by increasing the number of cameras attached to the rig. However, 360\textdegree~cameras have a problem of showing each other, so some ingenuity is required.

\section{Conclusion}

In this study, we propose a self-supervised learning method for depth estimation from a single 360\textdegree~ image. We introduce the coordinate transformation to cope with the geometric characteristics of Equirectangular Projection Format and the module to dynamically select features. 
Then, we conduct training and evaluation in an open outdoor environment, while existing work has been restricted to very limited environments. We use Multi-View Stereo to build a dataset for evaluation, and showed that the proposed method significantly improved the accuracy compared to existing methods.

{\small
\bibliographystyle{ieee_fullname}
\bibliography{egbib}
}



\clearpage

\noindent\scalebox{2.0}{{\bf Supplementary Material}}

\hfill

We write additional information -- analysis to the experimental results in Section A, details of Wang~\cite{Wang_2018_ACCV} and Wang+, in Section B, architectural details in Section C, and the datasets in Section D.


\hfill

\hfill

\noindent\scalebox{1.3}{{\bf A. Additional analysis of the experiments}}

\hfill

\noindent {\bf Why a dramatic improvement in Sq Rel for the proposed method over Wang~\cite{Wang_2018_ACCV} and Wang+ in Table \ref{table:result}? }
\ While our proposed method gives better results than Wang and Wang+, state-of-the-art, in all the evaluation metrics, there is a significant improvement, especially in Sq Rel. Sq Rel is the evaluation metric that weights the error,  with closer to the camera being more. As shown in Fig.\ref{fig:result}, our method shows better results in high-latitude areas which are corresponding to the bottom and top of the image. The lower part of the image (\ie, ground) tends to be closer, so more weight is given to it in Sq Rel. This is the reason why our method performs better than those methods in Sq Rel.

\hfill



\hfill


\noindent\scalebox{1.3}{{\bf B. Implementations of Wang and Wang+}}

\hfill

In this section, we give further information about the implementation details of Wang and Wang+.

\hfill

\noindent\scalebox{1.0}{{\bf Wang~\cite{Wang_2018_ACCV}}}. \ 
Since there is no publicly available source codes for this method, we implemented Wang~\cite{Wang_2018_ACCV} building upon the repository of SfM-Learner work~ \cite{Zhou_2017_CVPR} which is available online\footnote[1]{https://github.com/mrharicot/monodepth}.
We modified network architectures of depth and pose estimators including Cube Padding module by referring to the open code\footnote[2]{https://github.com/Yeh-yu-hsuan/BiFuse} of Wang~\etal~\cite{Wang_2020_CVPR} and Self Consistency Constraints by following Wang~\etal~\cite{Wang_2018_ACCV}.

\hfill

\noindent\scalebox{1.0}{{\bf Wang+}}. \ We implemented Wang+ as an extension of Wang~\cite{Wang_2018_ACCV} by using Godard~\etal~\cite{Godard_2019_ICCV} as the backbone architecture. 
Note that the implementation was based on publicly available code\footnote[3]{https://github.com/nianticlabs/monodepth2} but was extended to the 360\textdegree~imaginary. 
For Cube Padding, we substituted all the padding used in the encoder and decoder of Godard~\etal. 
We implemented Self Consistency Constraints as we did for Wang~\cite{Wang_2018_ACCV}. 
The models of Wang~\etal is optimized with Eq.\ref{equation:5}, where $\mathcal{L}_{rec}$ is reconstruction loss, $\mathcal{L}_{pose}$ is pose consistency loss of the six cubic faces, $\mathcal{L}_{sm}$ is smooth loss, and $\mathcal{L}_{exp}$ is explainability mask loss.
\begin{align}
\label{equation:5}
\mathcal{L}_{all} = \mathcal{L}_{rec} + \lambda_{pose}\mathcal{L}_{pose} + \lambda_{sm}\mathcal{L}_{sm} + \lambda_{exp}\mathcal{L}_{exp}
\end{align}
Except for $\mathcal{L}_{pose}$, we used the losses in Godard because we found they are better for self-supervised learning. 
For $\mathcal{L}_{pose}$, we tuned $\mathcal{\lambda}_{pose}$ and incorporated the optimal value into the loss.

\hfill

\noindent\scalebox{1.3}{{\bf C. Additional architectural details}}.

\hfill

\noindent\scalebox{1.0}{{\bf Overall network architecture}}.\ We extend the models proposed by Godard~\etal~\cite{Godard_2019_ICCV} by incorporating Distortion Aware Upsampling Module into decoder. Our depth estimation model uses ResNet50~\cite{He_2016_CVPR} as the encoder. The detailed architecture of the decoder part is shown in Table~\ref{table:architecture}. We use pose etimation model proposed by Godard, replacing Resnet18 with Resnet50. 

\hfill

\begin{table}[h!]
\scriptsize\centering

\begin{tabular}{lllllll}
                                    &            &            &               &              &                &                                          \\ \hline
\multicolumn{7}{|l|}{\textbf{Depth Decoder}}                                                                                                             \\ \hline
\multicolumn{1}{|l}{\textbf{layer}} & \textbf{k} & \textbf{s} & \textbf{chns} & \textbf{res} & \textbf{input} & \multicolumn{1}{l|}{\textbf{activation}} \\ \hline
\multicolumn{1}{|l}{upconv5}        & 3          & 1          & 256           & 32           & econv5         & \multicolumn{1}{l|}{ELU}                 \\
\multicolumn{1}{|l}{DAUM5}          & 1          & 1          & 256           & 16           & upconv5        & \multicolumn{1}{l|}{ELU}                 \\
\multicolumn{1}{|l}{iconv5}         & 3          & 1          & 256           & 16           & DAUM5, econv4  & \multicolumn{1}{l|}{ELU}                 \\ \hline
\multicolumn{1}{|l}{upconv4}        & 3          & 1          & 128           & 16           & iconv5         & \multicolumn{1}{l|}{ELU}                 \\
\multicolumn{1}{|l}{DAUM4}          & 1          & 1          & 128           & 8            & upconv4        & \multicolumn{1}{l|}{ELU}                 \\
\multicolumn{1}{|l}{iconv4}         & 3          & 1          & 128           & 8            & DAUM4, econv3  & \multicolumn{1}{l|}{ELU}                 \\
\multicolumn{1}{|l}{disp4}          & 3          & 1          & 1             & 1            & iconv4         & \multicolumn{1}{l|}{Sigmoid}             \\ \hline
\multicolumn{1}{|l}{upconv3}        & 3          & 1          & 64            & 8            & iconv4         & \multicolumn{1}{l|}{ELU}                 \\
\multicolumn{1}{|l}{DAUM3}          & 1          & 1          & 64            & 4            & upconv3        & \multicolumn{1}{l|}{ELU}                 \\
\multicolumn{1}{|l}{iconv3}         & 3          & 1          & 64            & 4            & DAUM3, econv2  & \multicolumn{1}{l|}{ELU}                 \\
\multicolumn{1}{|l}{disp3}          & 3          & 1          & 1             & 1            & iconv3         & \multicolumn{1}{l|}{Sigmoid}             \\ \hline
\multicolumn{1}{|l}{upconv2}        & 3          & 1          & 32            & 4            & iconv3         & \multicolumn{1}{l|}{ELU}                 \\
\multicolumn{1}{|l}{DAUM2}          & 1          & 1          & 32            & 2            & upconv2        & \multicolumn{1}{l|}{ELU}                 \\
\multicolumn{1}{|l}{iconv2}         & 3          & 1          & 32            & 2            & DAUM2, econv1  & \multicolumn{1}{l|}{ELU}                 \\
\multicolumn{1}{|l}{disp2}          & 3          & 1          & 1             & 1            & iconv2         & \multicolumn{1}{l|}{Sigmoid}             \\ \hline
\multicolumn{1}{|l}{upconv1}        & 3          & 1          & 8             & 2            & iconv2         & \multicolumn{1}{l|}{ELU}                 \\
\multicolumn{1}{|l}{DAUM1}          & 1          & 1          & 8             & 1            & upconv1        & \multicolumn{1}{l|}{ELU}                 \\
\multicolumn{1}{|l}{iconv1}         & 3          & 1          & 8             & 1            & DAUM1          & \multicolumn{1}{l|}{ELU}                 \\
\multicolumn{1}{|l}{disp1}          & 3          & 1          & 1             & 1            & iconv1         & \multicolumn{1}{l|}{Sigmoid}             \\ \hline
                                    &            &            &               &              &                &                                               
\end{tabular}
\caption{{\bf Network architecture}. Here {\bf k} is the kernel size, {\bf s} is the stride, and {\bf chns} is the number of output channels for each layer. Then, {\bf res } is the down-scaling factor for each layer relative to the input image. DAUM stands for Distortion Aware Upsampling Module.}
\label{table:architecture}
\end{table}

\begin{figure*}[h!]
\begin{center}
\includegraphics[width=\linewidth-10.cm,]{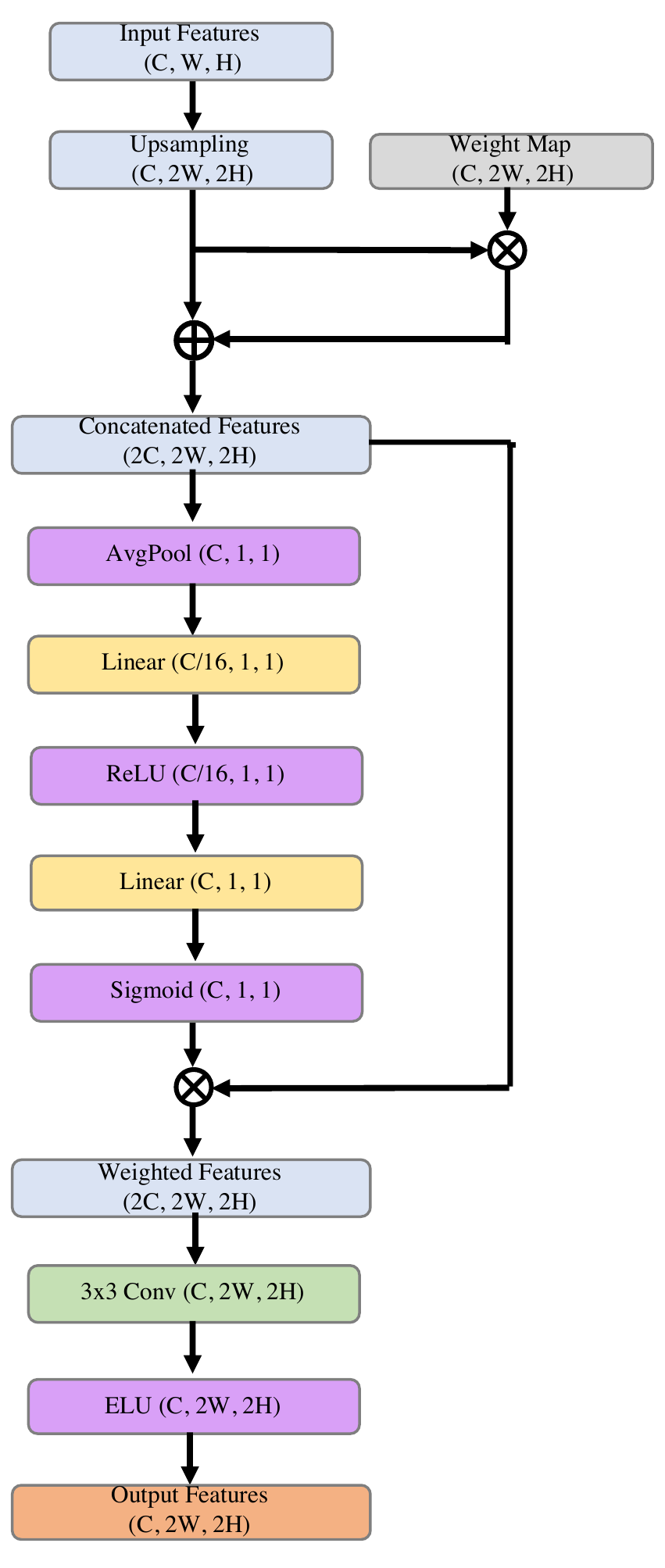}
\caption{{\bf The detailed architecture of the Distortion Aware Upsampling Module.} This is the details of each DAUM. For example, if this is DAUM4, then Input Features are upconv4, and C=128, W=(Original Image Width)/16 .}
\label{fig:DAUM_Arc}
\end{center}
\end{figure*}

\noindent\scalebox{1.0}{{\bf Distortion Aware Upsampling Module architecture}}.\  The detailed architecture of the Distortion Aware Upsampling Module is shown in Fig.~\ref{fig:DAUM_Arc}

\hfill

\noindent\scalebox{1.3}{{\bf D. Additional information about datasets}}

We used the publicly available source code of the Godard~\etal as a reference to determine the range of depths to be evaluated. Since Godard~\etal used 80 meters as the maximum depth for the KITTI dataset, we followed them and evaluated at distances up to 80 meters. 
Using the functions of the software\cite{metashape} for MVS, we reconstructed depth of scenes and used them as the ground truth data -- the average error of the reconstructed scene at each shooting spot was within a few centimeters.

\end{document}